\documentclass[letterpaper, 10 pt, conference]{IEEEtran}  

\usepackage{times}
\usepackage{graphicx}
\usepackage{epstopdf}

\usepackage{mdwmath}
\usepackage{mdwtab}
\usepackage{rotating}
\usepackage{caption}
\usepackage{subcaption}

\usepackage{amssymb}
\usepackage{amsfonts}
\usepackage{amsmath}
\usepackage{amsthm}
\usepackage{bm}

\usepackage{algorithm}
\usepackage{cite}

\usepackage{multirow} 
\usepackage{multicol}
\usepackage{array}

\usepackage{standalone}
\usepackage{booktabs} 

\usepackage{siunitx} 

\usepackage{accents}

\usepackage{pgfplotstable} 
\usepackage{verbatim}
\usepackage{isomath}
\usepackage{overpic}

\usepackage{tikz}
\usetikzlibrary{shapes,calc,patterns,
	decorations.pathmorphing,
	decorations.markings}

\tikzstyle{spring}=[very thick,decorate,decoration={zigzag,pre length=2,post
	length=2,segment length=6}]

\tikzstyle{damper}=[thick,decoration={markings, 
	mark connection node=dmp,
	mark=at position 0.5 with 
	{
		\node (dmp) [thick,inner sep=0pt,transform shape,rotate=-90,minimum
		width=15pt,minimum height=3pt,draw=none] {};
		\draw [thick] ( $(dmp.north east)+(2pt,0)$ ) -- (dmp.south east) -- (dmp.south
		west) -- ( $(dmp.north west)+(2pt,0)$ );
		\draw [thick] ( $(dmp.north)+(0,-5pt)$ ) -- ( $(dmp.north)+(0,5pt)$ );
	}
}, decorate]

\makeatletter\newcommand{\manuallabel}[2]{\def\@currentlabel{#2}\label{#1}}\makeatother

\usepackage{color}
\usepackage{xcolor}

\colorlet   {lightorange}{orange!20}
\colorlet   {lightgrey}  {gray!20}

\usepackage{amssymb}
\usepackage{mathtools}

\mathchardef\mhyphen="2D   

\newcommand{\RNum}[1]{\uppercase\expandafter{\romannumeral #1\relax}}

\graphicspath{
{figures/}
}

\providecommand{\figurename}{Fig.}

\usepackage[inline]{enumitem}
\setlist{nolistsep}

\usepackage[normalem]{ulem}                                        
\usepackage{marginnote}
\usepackage[textwidth=10ex,colorinlistoftodos]{todonotes}

\colorlet{fwu}{red}
\colorlet{ywu}{blue}
\colorlet{zbing}{green}

\usepackage{graphics} 
\usepackage{graphicx}
\usepackage{epsfig} 
\usepackage{times} 

\usepackage{amsmath, amssymb, amsthm}
\theoremstyle{definition} 

\usepackage{physics}
\usepackage{xcolor}
\usepackage{tikz}
\usepackage{graphicx}
\usepackage{booktabs}
\usepackage{amsmath}
\usepackage{algorithm}
\usepackage{algpseudocode}
\usepackage{tikz}
\usepackage{makecell}

\usetikzlibrary{calc} 
\usetikzlibrary{positioning}
\usetikzlibrary{tikzmark,decorations.pathreplacing}
\usepackage{balance}
\usepackage{soul}
\usetikzlibrary{arrows}
\usepackage{amsmath}
\usepackage{diagbox}
\usepackage[hidelinks]{hyperref}
\hypersetup{
    colorlinks=true,
    linkcolor=black,
    urlcolor=blue,
    citecolor=black
}
\usepackage[top=2.54cm, bottom=1.91cm, left=1.91cm, right=1.91cm]{geometry}

\IEEEoverridecommandlockouts

\title{\LARGE \bf%
React When You Need To: Event-Triggered Asynchronous Inference for VLA Policies }

\author{ Yansong Wu\textsuperscript{1*} , Huaqing Li\textsuperscript{1*} , Tianding Hou\textsuperscript{1*}, Lingyun Chen\textsuperscript{1,2}, Alois Knoll\textsuperscript{1} 
\thanks{
\textsuperscript{*} Equal contribution. $^{1}$ Technical University of Munich, Germany. $^{2}$ Mohamed Bin Zayed University of Artificial Intelligence, Abu Dhabi, UAE. Corresponding author: Yansong Wu ({\tt\small yansong.wu@tum.de}).

}%

}

\renewcommand{\baselinestretch}{0.98}

\let\oldtwocolumn\twocolumn
\renewcommand\twocolumn[1][]{%
    \oldtwocolumn[{#1}{
    \vspace{-15pt}
    \begin{center}
           \includegraphics[width=\textwidth]{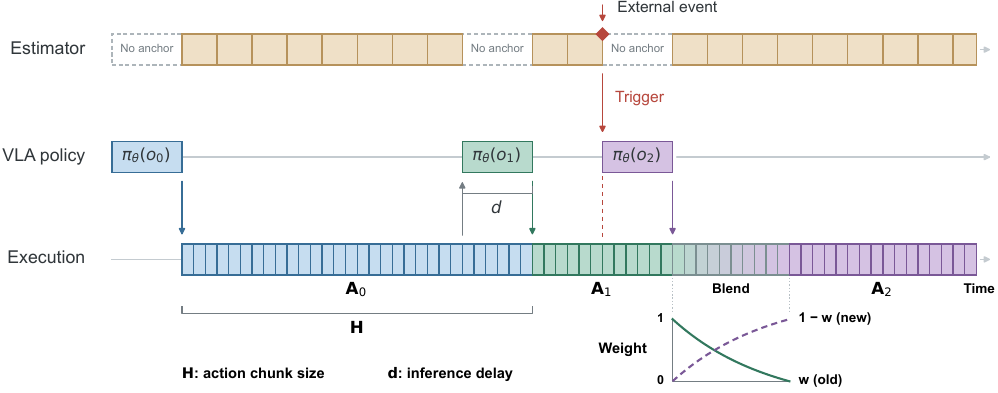}
           \captionof{figure}{Overview of the proposed event-triggered dynamic inference strategy. The event estimator continuously measures scene changes relative to the observation used for the most recent VLA inference. Without a significant event, the scheduler retains the maximum feasible inference gap to preserve motion consistency. Upon event detection, it triggers early inference and smoothly transitions to the updated action chunk for prompt reactivity.}
            \label{fig:policy-execution-trigger}
    \end{center}
    \vspace{5pt}
    }]
}

\begin{document}
\maketitle

\begin{abstract}
Vision-Language-Action (VLA) models commonly predict action chunks, limiting their ability to react to environmental changes during execution. Existing asynchronous inference methods improve reactivity but typically rely on a fixed inference gap. In this paper, we propose an event-guided dynamic inference strategy that adapts the inference gap according to scene changes observed since the previous inference. Thereby, it simultaneously preserves motion consistency and prompt reactivity. Across static and dynamic real-world settings, our method consistently performs best, averaging \(95\%\) success and exceeding the strongest baseline by \(55\) percentage points. The code will be made publicly available upon acceptance. The project page is available at \href{https://react-when-you-need-to.github.io/}{this URL}.

\end{abstract}

\section{Introduction}

Vision-Language-Action (VLA) models have shown strong generalization across robot manipulation tasks~\cite{brohan2022rt1,zitkovich2023rt2,driess2023palme,octo2024,kim2024openvla,kim2025finetuning}.  Existing VLA models commonly predict action chunks, generating multiple future actions per inference~\cite{zhao2023learning}. Under conventional synchronous execution, the robot completes the current chunk before requesting the next one. Consequently, when unexpected environmental changes occur during chunk execution, the robot continues to follow the precomputed action chunk and cannot react until the next policy inference.

Although primarily developed to eliminate inter-chunk stalls, asynchronous execution also partially mitigates this delayed response~\cite{black2024rtc,shukor2025smolvla,sendai2025a2c2,tang2025vlash,black2025training,xie2026dynamicvla,yang2026abpolicy}. It initiates the next policy inference while the robot continues executing the ongoing action chunk. It enables more frequent policy updates, thereby reducing the response latency to environmental changes.

However, existing asynchronous execution methods typically use a fixed inference gap, creating an inherent trade-off between reactivity and execution reliability. Shortening the inference gap enables more frequent policy inference and faster reactions to environmental changes. However, it also necessitates more frequent merging of consecutive action chunks, which may introduce trajectory inconsistencies and compromise execution reliability~\cite{chi2025diffusion,liu2408bidirectional}. In contrast, a longer inference gap better preserves motion continuity but delays the robot's response to environmental changes~\cite{black2024rtc}.

To address this limitation, we propose an event-triggered dynamic inference strategy that schedules VLA inference according to environmental changes. During execution, an event estimator measures the degree of change between the current scene and the anchor scene observed during the most recent policy. The resulting event signal indicates whether early policy inference is required. While maintaining stall-free execution, our strategy retains the largest feasible inference gap when no significant event is detected. Once an event is detected, it immediately triggers a new inference without waiting until the default gap. In this way, our strategy preserves the motion consistency of long inference gaps while retaining the prompt reactivity of short ones.

The main contributions are summarized as follows:
\begin{itemize}
    \item It maintains the largest feasible gap in the absence of significant events and triggers early inference upon event detection, thereby simultaneously preserving motion consistency and prompt reactivity.

    \item We develop a lightweight event estimator based on the VLA vision encoder to quantify scene changes since the most recent policy inference. Our event estimation is interleaved with VLA inference and operates at up to $15~\mathrm{Hz}$ in our setup.

\item We conduct comprehensive real-world experiments involving both static and dynamic tasks. Our method achieves an average full-task success rate of \(95\%\) across the three evaluation settings, representing an absolute improvement of \(55\)\% over the best competing methods.
\end{itemize}

\section{Related Work}
\subsection{Vision-Language-Action Models}
Vision-Language-Action (VLA) models~\cite{brohan2022rt1,zitkovich2023rt2,driess2023palme,octo2024,kim2024openvla,kim2025finetuning} leverage large-scale vision-language pretraining to learn end-to-end policies that map visual observations and language instructions to low-level robot actions. Recent advances have followed several complementary architectural directions. Autoregressive models~\cite{kim2024openvla,kim2025finetuning,pertsch2025fast} formulate robot control as next-token prediction over discretized action tokens, naturally extending the training paradigm of vision-language models to action generation. Diffusion-based approaches~\cite{chi2023diffusionpolicy,guo2024prediction,octo2024,liu2025hybridvla,zhou2025chatvla} instead model continuous action distributions, enabling smooth and expressive behaviors at the cost of iterative inference. Flow-based methods, exemplified by $\pi_0$~\cite{black2024pi0}, replace diffusion sampling with conditional flow matching to improve inference efficiency. Meanwhile, dual-system architectures such as $\pi_{0.5}$~\cite{black2025pi05} and GR00T-N1~\cite{nvidia2025gr00tn1} decouple high-level semantic reasoning from low-level action generation to support long-horizon manipulation. Nevertheless, VLA models are commonly deployed with synchronous inference, where action generation and robot execution proceed sequentially. This execution strategy introduces pronounced stalls between consecutive action chunks and limits the policy's ability to respond to environmental changes during chunk execution~\cite{black2024rtc}.

\subsection{Asynchronous Inference}

To eliminate this stop-and-go behavior, asynchronous inference has received increasing research attention~\cite{black2024rtc,shukor2025smolvla,sendai2025a2c2,tang2025vlash,black2025training,xie2026dynamicvla,yang2026abpolicy}. It generates the next action chunk while the current one is still being executed, ensuring continuous action availability without waiting for model inference. SmolVLA~\cite{shukor2025smolvla} adopts such an asynchronous execution scheme, but directly switching between independently generated chunks can introduce inter-chunk discontinuities. RTC~\cite{black2024rtc} mitigates this issue by freezing the action prefix guaranteed to be executed during inference and inpainting the remaining actions, while softly incorporating additional overlapping actions to improve continuity. However, its gradient-based inference-time guidance introduces additional computational overhead. Training-time RTC~\cite{black2025training} addresses this limitation by simulating inference delays during training and directly learning to generate an action postfix conditioned on the committed prefix, thereby eliminating the additional inference-time guidance. Other studies address the temporal misalignment between prediction and execution through execution-time state conditioning~\cite{tang2025vlash}, outdated-action filtering~\cite{xie2026dynamicvla}, per-step action correction~\cite{sendai2025a2c2}, or trajectory refitting~\cite{yang2026abpolicy}. Despite these advances, existing asynchronous methods typically invoke the policy continuously or at a predefined rate, without adapting the inference schedule to environmental changes. 




\section{Problem Formulation}
Following the definition in~\cite{black2024rtc}, an action chunking policy is denoted by $\pi(\mathbf{A}_t | \mathbf{o}_t)$, where $\mathbf{A}_t = [\mathbf{a}_t, \mathbf{a}_{t+1}, \dots, \mathbf{a}_{t+H-1}]$ is a chunk of future actions, $\mathbf{o}_t$ is an observation, $t$ indicates a controller timestep, and $H$ denotes the \textit{prediction horizon}. 
Let $\Delta t$ be the sampling period of the robot controller, and let $\delta$ be the continuous time required for the policy to generate an action chunk. The \textit{inference delay} is defined as $d := \lfloor \delta / \Delta t \rfloor$. The execution horizon $s$ denotes the number of actions actually executed from each chunk.

On this basis, let $\Delta T$ denote the elapsed time between the completion of one inference call and the initiation of the next. We define the corresponding number of controller steps as the \textit{inference gap} $g$:\vspace{-4pt}
\begin{equation}
g := \left\lfloor \frac{\Delta T}{\Delta t} \right\rfloor.\label{eq:gap}
\end{equation}
The robot can operate continuously without inter-chunk execution stalls, as long as the system parameters satisfy: \begin{equation}
 g \le H -d.
\end{equation}
In practice, selecting an appropriate inference gap $g$ presents a fundamental trade-off. A small gap enables frequent policy updates and improves responsiveness, but also leads to frequent transitions between action chunks, which may introduce action instable execution~\cite{chi2025diffusion, liu2408bidirectional}. Conversely, a large gap reduces the frequency of chunk transitions but makes the policy less responsive to environmental changes occurring during chunk execution~\cite{black2024rtc}. 

This trade-off motivates our central problem: \textit{can the inference gap be dynamically adapted to environmental changes while preserving seamless execution?} Specifically, the system is expected to maintain a larger inference gap when little environmental change has occurred since the current action chunk was predicted, and automatically switch to a small gap when a significant change is detected, enabling the policy to respond with minimal delay.

\section{Method}
\label{sec:method}

To address this problem, we propose an event-triggered dynamic inference framework that uses a data-driven event estimator to monitor whether the currently executing action chunk remains applicable to the evolving scene and schedules policy inference accordingly.

\subsection{Event-Supervision Construction from Human Demonstrations}
Rather than manually annotating the occurrence and relevance of such events, we automatically derive event supervision from human demonstrations. The entire process consists of three steps: 


\subsubsection{Data Collection}
For each task, we collect 40 demonstrations under static environmental conditions and 160 demonstrations under dynamic environmental conditions via teleoperation, forming \(\mathcal{D}_{\mathrm{static}}\) and \(\mathcal{D}_{\mathrm{dynamic}}\), respectively.
The complete demonstration dataset is formed by combining the two:
\begin{equation}
    \mathcal{D}
    =
    \mathcal{D}_{\mathrm{static}}
    \cup
    \mathcal{D}_{\mathrm{dynamic}}.
\end{equation}
The dataset comprises $N$ demonstrations:
\begin{equation}
    \mathcal{D} = \left\{ \tau^{(i)} \right\}_{i=1}^{N},
\end{equation}
where each demonstration $\tau^{(i)}$ is defined as:
\begin{equation}
    \tau^{(i)} = \left( \ell^{(i)}, \big\{ (\bm{I}_t^{(i)}, \bm{s}_t^{(i)}, \bm{a}_t^{(i)}) \big\}_{t=1}^{T_i} \right).
\end{equation}
Here, $i \in \{1, \dots, N\}$ indexes the demonstration, and $T_i$ denotes its trajectory length. Specifically, $\ell^{(i)}$ represents the language instruction, while $\bm{I}_t^{(i)}$, $\bm{s}_t^{(i)}$, and $\bm{a}_t^{(i)}$ denote the visual image, robot proprioceptive state, and action at time step $t$, respectively.

\subsubsection{VLA Fine-Tuning}
We fine-tune the pretrained $\pi_{0.5}$ VLA policy on $\mathcal{D}_{\mathrm{static}}$ using LoRA~\cite{hu2022lora} to establish a reference model of nominal task behavior. The dynamic subset is excluded because actions recorded after an abrupt event reflect the demonstrator's adaptation to environmental information unavailable in the initial observation. Including such actions would not only violate the causal alignment between the initial observation and its supervised action chunk, but also incorporate event-induced responses into the nominal reference.

After fine-tuning, all VLA parameters are frozen. The resulting reference policy is subsequently applied offline to the complete demonstration dataset to generate nominal action chunks for event-supervision construction. 

\subsubsection{Event Estimator Dataset Construction}
\begin{algorithm}
\caption{Dataset Construction for Event Estimator}
\label{alg:event-data-generation}
\begin{algorithmic}[1]
\Require VLA model $\pi$,  Dataset $\mathcal{D}$, vision encoder $E_\mathrm{vis}$, horizon $H$, and inference latency $d$
\State Initialize $\mathcal{D}_\mathrm{event} \gets \emptyset$

\For{$i = 1$ \textbf{to} $N$}
    \For{$j = 1$ \textbf{to} $T_i - H+1$}
        \State $\bm{A}_j^{(i)} \gets [\bm{a}_{j}^{(i)}, \dots, \bm{a}_{j+H-1}^{(i)}]$ \Comment{demo actions}
        \State $\hat{\bm{A}_j}^{(i)} \gets \pi\big( \ell^{(i)}, \bm{I}_j^{(i)}, \bm{s}_j^{(i)} \big)$ \Comment{predicted actions}
        \State $\bm{z}_{anchor}^{(i)} \gets E_\mathrm{vis}(\bm{I}_j^{(i)})$ \Comment{anchor visual feature}
        \For{$k = 1$ \textbf{to} $H-d$}
            \State $\bm{z}_{j+k}^{(i)} \gets E_\mathrm{vis}(\bm{I}_{j+k}^{(i)})$ \Comment{current visual feature}
            \State $e^{(i)}_{j, j+k} \gets \operatorname{MSE}(\bm{A}_j^{(i)}[:k], \, \hat{\bm{A}_j}^{(i)}[:k])$ \label{alg:line-pair}
            \State $\bm{p} \gets \big( \bm{z}_{anchor}^{(i)}, \bm{z}_{j+k}^{(i)}, \, e^{(i)}_{j, j+k} \big)$ \Comment{event pair}
            \State $\mathcal{D}_\mathrm{event} \gets \mathcal{D}_\mathrm{event} \cup \{ \bm{p} \}$
        \EndFor
    \EndFor
\EndFor
\State \Return $\mathcal{D}_\mathrm{event}$
\end{algorithmic}
\end{algorithm}

Afterwards, we leverage the fine-tuned VLA policy to perform offline inference across the demonstration dataset $\mathcal{D}$. As detailed in Algorithm~\ref{alg:event-data-generation}, for each inference performed, we record the corresponding visual representation as the anchor visual feature $\bm{z}_{anchor}^{(i)}$. We then compare the predicted chunk with the corresponding demonstrated actions, whose divergence indicates whether an event occurring during chunk execution has altered the intended behavior. Specifically, we define the event score $e_{j,j+k}$ as a continuous proxy for the likelihood that an abrupt event has occurred during chunk execution and altered the intended motion (Line 9 in Algorithm~\ref{alg:event-data-generation}). The first subscript of event score denotes the anchor inference timestep, and the second identifies the current time step evaluated against that anchor. Collecting the labeled anchor-current feature pairs yields the event estimator dataset $\mathcal{D}_{\mathrm{event}}$.

\subsection{Event Estimator}
\subsubsection{Event Score Estimation}
Based on the event dataset $\mathcal{D}_{\mathrm{event}}$ constructed above, we develop a lightweight event estimator that predicts the event score by comparing the visual observation captured at the most recent policy inference with the current observation.

Specifically, our event estimator consists of a frozen VLA vision encoder $E_\mathrm{vis}$ and a lightweight Transformer-based event decoder $D_\mathrm{event}$. Let $t_0$ denote the time step of the most recent policy inference (anchor) and $t$ denote the current time step ($t > t_0$). The vision encoder extracts visual representations $\bm{z}_{t_0}$ and $\bm{z}_t$ from the respective observations $\bm{I}_{t_0}$ and $\bm{I}_t$.
\begin{equation}
\begin{aligned}
    \bm{z}_{t_0} &= E_{\mathrm{vis}}(\bm{I}_{t_0}), \\
    \bm{z}_{t}   &= E_{\mathrm{vis}}(\bm{I}_{t}).
\end{aligned}
\label{eq:paired-visual-features}
\end{equation}
The transformer-based event decoder then compares the resulting visual representations and predicts the corresponding event score $e_{t_0,t}$:
\begin{equation}
    e_{t_0,t}
    = D_{\mathrm{event}}(\bm{z}_{t_0},\bm{z}_t),
    \label{eq:event-score}
\end{equation}
where the first subscript of event score denotes the anchor scene, and the second denotes the current scene compared against it.

Reusing the pretrained VLA visual encoder instead of introducing another backbone offers two key advantages: (i) \textit{Task-Aligned Representations:} Pretrained on manipulation scenes to support action generation, the encoder inherently preserves task-critical spatial and semantic information. (ii) \textit{Computational Efficiency: }The anchor feature $\bm{z}_{t_0}$ is already computed during policy inference and can be cached at zero marginal cost. Furthermore, extracting $\bm{z}_t$ and evaluating $D_\mathrm{event}$ run within the interval between policy inferences, avoiding introducing computational latency to action generation.  

\subsubsection{Score-to-Event Mapping}
\begin{figure}[h]
    \centering
    \includegraphics[width=0.8\linewidth]
{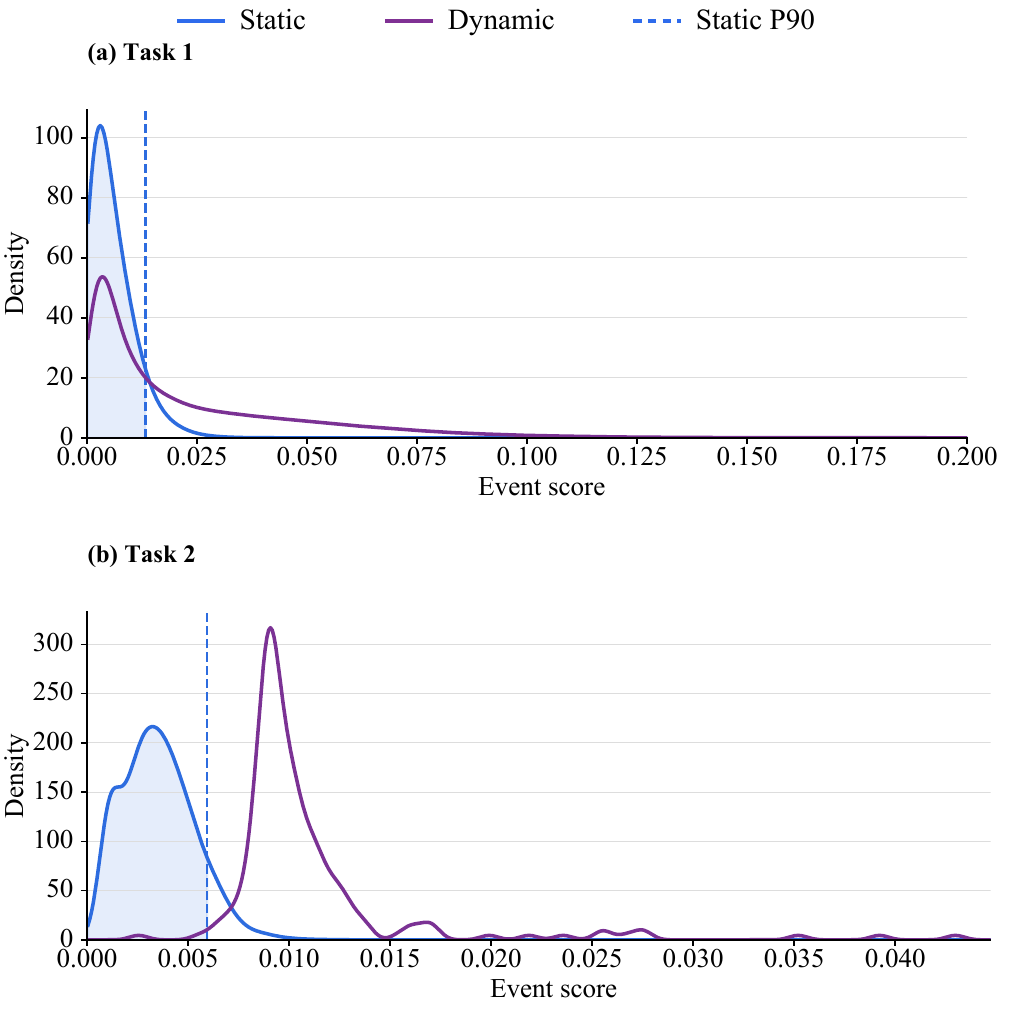}
   \caption{Representative event-score distributions under static and dynamic conditions for Task~1 and Task~2 detailed in Sec.~\ref{sec:task}. The dashed blue line marks the task-specific 90th percentile of the static distribution, used as the event threshold \(\theta_{\mathrm{P90}}\).}
    \label{fig:event-score-distributions}
\end{figure}

To convert the continuous event score into a binary inference decision, we define a P90 threshold $\theta_{P90}$ as the \textit{90th} percentile of the event-score distribution obtained from static demonstrations. As illustrated in Fig.~\ref{fig:event-score-distributions}, although the event-score distributions of dynamic demonstrations vary substantially across tasks, the static distributions exhibit similar shapes and consistently concentrate in the low-score region. Scores from static demonstrations primarily characterize variations that should not trigger an early policy inference. We therefore treat the lower 90\% of the static distribution as non-event variation. To suppress triggers caused by transient score fluctuations, an event is declared only when the estimated score exceeds $\theta_\mathrm{P90}$ in two consecutive evaluations, upon which a new policy inference is triggered. 

\subsection{Event-Driven Inference Scheduling}
As illustrated in Fig.~\ref{fig:policy-execution-trigger}, policy inference and event estimation run alternately without temporal overlap, thereby avoiding competition for computational resources. Each rollout begins with an initial policy inference that generates an action chunk from the current observation. The visual feature extracted from the inference-time scene image by the VLA vision encoder is retained as the anchor feature for subsequent event estimation. Once policy inference is complete, the event estimator runs repeatedly, comparing the latest available observation with the anchor to determine whether an event requiring a new policy inference has occurred.

If no event is detected, the next policy inference is automatically initiated after $H-d$ actions of the current chunk have been executed, leaving the final $d$ actions to cover the inference delay and ensure seamless execution. If an event is detected before this point, the system instead triggers the next inference immediately using the latest observation.

For this event-triggered early inference, the new action chunk becomes available before the ongoing chunk has been fully executed. To ensure a smooth transition, we blend the overlapping actions as follows:



For an overlap of $N \geq 2$ actions, the blending
weights are defined as
\begin{equation}
\mathbf w =
\frac{
\exp\!\left(-\beta\frac{\mathbf k}{N-1}\right)
-\exp(-\beta)\mathbf 1_N
}{
1-\exp(-\beta)
},
\label{eq:merge_weight}
\end{equation}
where $\mathbf k=[0,1,\ldots,N-1]^{\top}$ indexes the
actions in the overlapping area, $N$ is the overlap
length,$\mathbf{1}_N\in\mathbb{R}^N$ is the all-ones vector, and $\beta>0$ controls the decay of the weights.
The exponential is applied element-wise.
The weight assigned to the previous chunk decreases
from $1$ to $0$.

The overlapping actions are then merged as:
\begin{equation}
\mathbf A_{\mathrm{merge}}
=
\mathbf w \odot \mathbf A_{\mathrm{old}}
+
(\mathbf 1-\mathbf w)\odot \mathbf A_{\mathrm{new}},
\label{eq:chunk_merge}
\end{equation}
where $\mathbf A_{\mathrm{old}}$ and $\mathbf A_{\mathrm{new}}$ denote the overlapping segments of the previous and new chunks, respectively.


\begin{figure*}[t]
    \centering

    \begin{subfigure}{\textwidth}
        \centering
        \includegraphics[width=\linewidth]
        {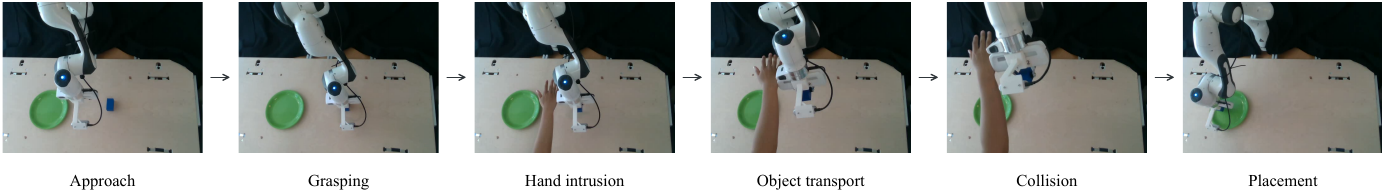}
        \caption{Synchronous execution resulting in a collision.}
        \label{fig:process-normal}
    \end{subfigure}

    \vspace{4mm}

    \begin{subfigure}{\textwidth}
        \centering
        \includegraphics[width=\linewidth]
        {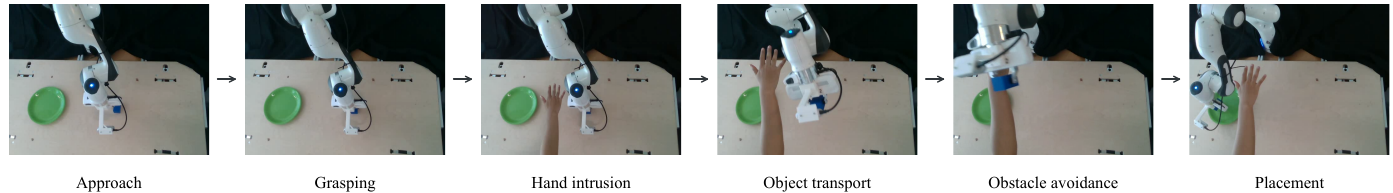}
        \caption{Our method successfully avoiding the dynamic obstacle.}
        \label{fig:process-ours}
    \end{subfigure}
    \vspace{-1mm}

    \caption{Qualitative comparison of synchronous execution and our method under repeated obstructions caused by a moving human hand during a pick-and-place task.}
    \label{fig:execution-comparison}
\end{figure*}

\section{Experiment}
\label{sec:experiment}

In this section, we conduct two groups of real-world manipulation experiments to address the following research questions: 

\begin{itemize}
    \item \textbf{Q1:} Can the proposed method maintain  reliable task execution in static environments?

    \item \textbf{Q2:} Can the proposed method respond promptly and effectively to environmental changes through event-triggered inference scheduling?
    \item \textbf{Q3:} Can our method achieve consistently high task performance across static and dynamic settings?

\end{itemize}


\subsection{Experimental Setup}

\begin{table}[h]
\centering
\setlength{\tabcolsep}{5pt}
\caption{Configuration Details}
\label{tab:runtime}

\begin{tabular}{@{}lcc@{}}
\toprule
\textbf{Setting} & & \textbf{Value} \\
\midrule
VLA fine-tune steps            & & $30{,}000$ \\
VLA fine-tune batch size       & & $32$ \\
Action-chunk horizon $H$       & & $50$ \\
Blending weight decay $\beta$  & & $1.2$ \\
\midrule
Demonstration sampling rate    & & $30\,\mathrm{Hz}$ \\
Execution frequency           & & $30\,\mathrm{Hz}$ \\
Event estimator frequency         &   & $15\,\mathrm{Hz}$ \\
\midrule
Inference delay $d$  & & $5$ \\
RTC inference delay $d_{\mathrm{RTC}}$ & & $8$ \\
RTC inference gap $g$           &   & $20$  \\
DynamicVLA inference gap  $g$     &   & $0$ \\
\bottomrule
\end{tabular}
\par\vspace{4pt}
\begin{minipage}{\columnwidth}
\footnotesize
\raggedright
\textsuperscript{\(\dagger\)}
The inference delay is $d=5$ action steps for all methods
except RTC. RTC uses an inference delay of
$d_{\mathrm{RTC}}=8$ action steps, corresponding to an
execution horizon of $s=28$ in the RTC formulation.
\end{minipage}
\vspace{-4pt}
\end{table}

\subsubsection{Robotic Platform}
The experimental platform consists of a Franka Emika Panda robot and three Intel RealSense D435i cameras. The robot controller runs on a NUC with an Intel i7-10700 CPU, while the VLA model and the proposed event estimator are executed on a separate workstation equipped with an NVIDIA RTX 3090 GPU. We fine-tune the VLA model using six NVIDIA A40 GPUs.  Table~\ref{tab:runtime} summarizes the concrete training and inference configurations
\subsubsection{Evaluation Tasks \label{sec:task}}
We evaluate the proposed method on two manipulation tasks involving distinct types of environmental changes.

\begin{itemize}
    \item \textbf{Task~1: Pick-and-place with static and dynamic obstacles }(Fig.~\ref{fig:process-ours}). The robot is instructed to pick up a designated blue cube and place it on a green plate while avoiding an obstacle in the transport region. The obstacle is introduced by a human operator extending a hand into the workspace. We evaluate two conditions: (i) In the static condition, the obstacle is introduced before transportation and remains stationary, while its location is varied across trials. (ii) In the dynamic condition, we adopt a closed-loop obstruction protocol. During transport, the operator introduces the hand into the robot's current motion path without intentionally making contact. Whenever the robot changes its path to avoid the hand, the operator repositions it to obstruct the updated path. This process continues until the robot reaches the placement region. 

    \item \textbf{Task~2: Placement onto a moving target} (Fig.~\ref{fig:task2-process}). The robot starts with an object already grasped and is instructed to place it on a movable plate. We adopt a closed-loop target-relocation protocol in which the operator repeatedly moves the plate during execution. Whenever the grasped object approaches the region above the plate, the operator moves the plate away from the robot's current placement trajectory, varying both its position and direction of motion. This process continues until the robot successfully places the object or the trial terminates. Consequently, successful execution requires the policy to repeatedly update its ongoing action sequence based on recent observations and track the relocated target. 
\end{itemize}

\subsection{Comparing Methods}
We compare our method against three representative execution baselines and four fixed-gap ablations. All configurations use the same $\pi_{0.5}$ policy backbone~\cite{black2025pi05}.
\begin{itemize}
\item \textbf{Synchronous:} Policy inference and action execution alternate without overlap.
\item \textbf{DynamicVLA:} DynamicVLA~\cite{xie2026dynamicvla} improves responsiveness in dynamic environments by discarding actions whose intended execution steps elapse during inference and updating the execution buffer with the remaining chunk. For a fair comparison, we replace its original VLA backbone with $\pi_{0.5}$ while retaining this update strategy.
\item \textbf{RTC:} RTC~\cite{black2024rtc} uses inference-time inpainting and soft masking to condition new action chunks on previously committed actions.

\item \textbf{Ours:} We apply the proposed method described in Section~\ref{sec:method} to the shared backbone.

\item \textbf{Fixed-gap triggering ablations:} 
We disable the event estimator and use four fixed inference gaps, $g\in\{0,5,25,45\}$, to schedule subsequent inferences. Each variant is denoted as Ablation ($g=x$), where $x$ is the corresponding fixed gap. All other components and execution settings remain unchanged.

\end{itemize}

\subsection{Experimental Procedure}

Each method is evaluated in 20 full-task trials under each of the three settings: Task~1 with a static obstacle, Task~1 with a dynamic obstacle, and Task~2 with a moving target. Since Task~1 comprises three sequential stages, a failure at an earlier stage would otherwise prevent the evaluation of the remaining stages. We manually restore the required intermediate state and resume execution, ensuring that each stage is evaluated over 20 attempts. Such recovery contributes only to the stage-wise evaluation, while the original full-task trial remains recorded as a failure.

\begin{figure*}[t]
    \centering
    \includegraphics[width=0.8\textwidth]{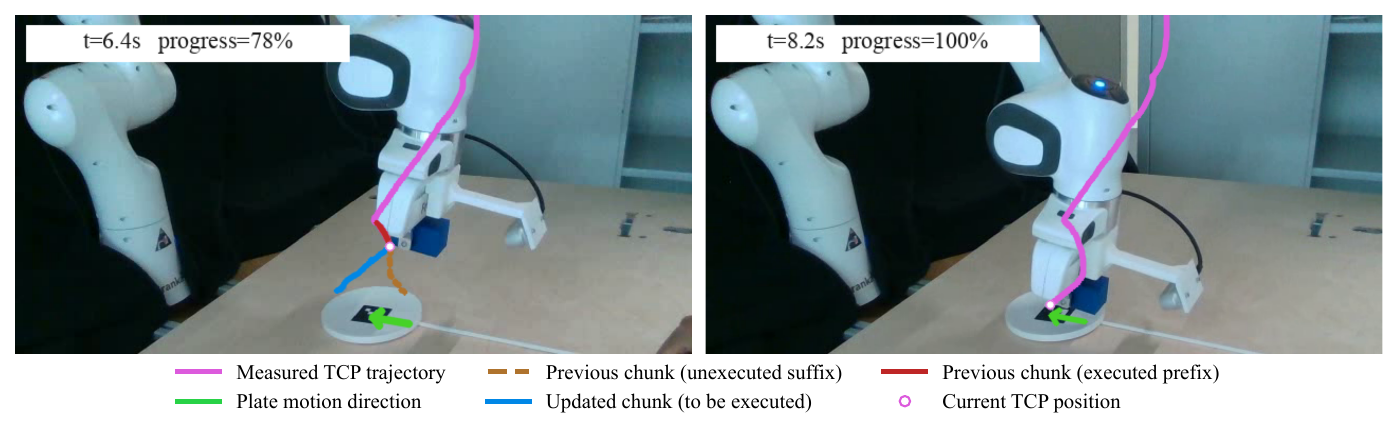}
    \caption{Representative Task 2 execution under target-motion perturbation, showing the action-chunk update at \(t=6.4\,\mathrm{s}\) and the complete TCP trajectory at task completion (\(t=8.2\,\mathrm{s}\)).}
    \label{fig:task2-process}
\end{figure*}

\subsection{Evaluation Metrics}
For \textit{Task~1}, under both static and dynamic obstacle conditions, performance is evaluated at the stage and full-task levels. A pick is successful when the robot grasps the designated cube; obstacle avoidance is successful when the robot transports the grasped cube to the placement region without contacting the obstacle; and placement is successful when the robot releases the cube onto the target plate. A full-task success requires all three stages to be completed successfully within a single trial. \textit{Task~2} evaluates only the placement on moving target and adopts the same success criterion, requiring the grasped object to be released onto the moving plate.

Based on the above criteria, the success rate for each stage or full task is computed as:
\begin{equation}
    \mathrm{SR}_{c}
    =
    \frac{N_{c}^{\mathrm{success}}}{N_{c}^{\mathrm{trials}}}
    \times 100\%,
    \label{eq:success_rate}
\end{equation}
where $c$ denotes the evaluated stage or full task, and
$N_{c}^{\mathrm{success}}$ and $N_{c}^{\mathrm{trials}}$ denote the corresponding numbers of successful and total trials, respectively.


\section{Results and Discussion}

\subsection{Performance in Static Environments (Q1)}

Table~\ref{tab:stage-breakdown-baselines} reports the phase-wise and full-task success rates of the evaluated execution methods in the pick-and-place task with static obstacles. All methods performed well in the obstacle-avoidance and placement phases, achieving 100\% success in both phases, except for DynamicVLA that achieved a placement success rate of 90\%. The performance variance primarily concentrated  in the picking phase. Synchronous execution, the fixed-gap ablations ($g=25$ and $g=45$), and our method achieved 100\% picking success, whereas the ablation with $g=5$ achieved only 30\%. By contrast, DynamicVLA, RTC, and the ablation with $g=0$ failed in all picking trials. Compared with the other two phases, picking places higher demands on precise gripper--object alignment. Inspection of the failed trials showed that most failures occurred when the gripper tip pressed against the object during grasping, displacing it before a stable grasp could be established.

\begin{table}[h]
\centering

\caption{Stage-wise and full-task success rates for Task~1.}
\label{tab:stage-breakdown}
\vspace{6pt}

\begin{subtable}[t]{\columnwidth}
\centering
\caption{Success Rate with Static Obstacle [\%]}

\label{tab:stage-breakdown-baselines}
\vspace{-4pt}
\resizebox{\linewidth}{!}{%
\begin{tabular}{lcccc}
\toprule
\textbf{Mode} &
\textbf{Pick} &
\textbf{Obstacle Avoidance} &
\textbf{Place} &
\textbf{Full-Task} \\
\midrule
Synchronous      & 100 & 100 & 100 & 100 \\
DynamicVLA       &   0 & 100 &  90 &   0 \\
RTC              &   0 & 100 & 100 &   0 \\
\midrule
Ablation ($g=0$)  &   0 & 100 & 100 &   0 \\
Ablation ($g=5$)  &  30 & 100 & 100 &  30 \\
Ablation ($g=25$) & 100 & 100 & 100 & 100 \\
Ablation ($g=45$) & 100 & 100 & 100 & 100 \\
\midrule
Ours             & 100 & 100 & 100 & 100 \\
\bottomrule
\end{tabular}%
}
\end{subtable}

\par\vspace{12pt}

\begin{subtable}[t]{\columnwidth}
\centering
\caption{Success Rate with Dynamic Obstacle [\%]}

\label{tab:stage-breakdown-ablations}
\vspace{-4pt}
\resizebox{\linewidth}{!}{%
\begin{tabular}{lcccc}
\toprule
\textbf{Mode} &
\textbf{Pick} &
\textbf{Obstacle Avoidance} &
\textbf{Place} &
\textbf{Full-Task} \\
\midrule
Synchronous      & 100 &   0 & 100 &   0 \\
DynamicVLA       &   0 &   0 & 100 &   0 \\
RTC              &  15 &  95 & 100 &  15 \\
\midrule
Ablation ($g=0$)  &   0 &  95 & 100 &   0 \\
Ablation ($g=5$)  &  15 &  90 & 100 &  15 \\
Ablation ($g=25$) &  90 &   0 &  95 &   0 \\
Ablation ($g=45$) & 100 &   0 & 100 &   0 \\
\midrule
Ours             & 100 & 100 & 100 & 100 \\
\bottomrule
\end{tabular}%
}
\end{subtable}

\par\vspace{6pt}
\begin{minipage}{\columnwidth}
\footnotesize
\raggedright
\textsuperscript{\(\dagger\)}
All stage metrics report success rates across 20 trials, with manual resets upon failure in prior stages. Full-Task indicates end-to-end success without intervention.
\end{minipage}
\vspace{-8pt}
\end{table}

The fixed-gap ablation tests reveal a clear trend: picking success increased markedly as the inference gap increased. In static environments, short inference gaps lead to frequent action-chunk merging, which may disrupt the fine-grained motions required for grasping. In contrast, longer gaps allow longer action sequences to be executed without interruption, thereby preserving motion consistency and improving grasping reliability. By maintaining a long inference gap when no significant environmental change is detected, our method achieves a 100\% picking success rate, matching the fixed long-gap ablations.

\subsection{Performance in Dynamic Environments (Q2)}
\subsubsection{Pick-and-Place with Dynamic Obstacle}
As shown in Table~\ref{tab:stage-breakdown-ablations}, interaction with the moving obstacle occurs primarily during the transport phase after the object has been grasped. Consequently, the results for the picking and placement phases remain broadly consistent with those observed in the static setting. During dynamic obstacle avoidance, our method again achieves the highest success rate of 100\%. However, the baseline methods exhibit a trend almost opposite to that observed in the static setting. Methods with frequent policy updates, including RTC and the short-gap ablations ($g=0$ and $g=5$), perform strongly in dynamic obstacle avoidance, whereas synchronous execution and the long-gap ablations ($g=25$ and $g=45$) fail to react to the moving obstacle.

\begin{figure}
    \centering
    \includegraphics[width=\linewidth]{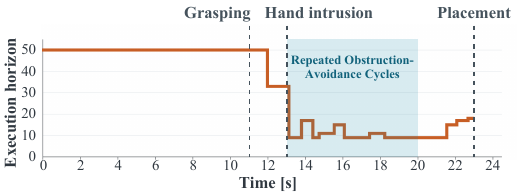}
    \caption{Execution horizon during Task~1.
Dashed lines mark task events. The shaded region indicates repeated obstruction-avoidance cycles.}
    \label{fig:executed_actions_t1}
\end{figure}

An illustrative comparison is shown in Fig.~\ref{fig:execution-comparison}. Upon a sudden hand intrusion, our event estimator promptly detects the environmental change and triggers an early policy inference. As the person repeatedly moves the hand to block the robot, the resulting event triggers maintain a high inference frequency, allowing the robot to continuously update its action plan and maneuver around the hand. Figure~\ref{fig:executed_actions_t1} visualizes the number of actions executed from each predicted action chunk during this process. Following the hand intrusion, the executed chunk lengths decrease substantially, directly reflecting the repeated early-inference triggers required to respond to the continuously changing obstacle. In contrast, synchronous execution cannot respond to unexpected changes that occur during the execution of an action chunk. It therefore continues executing the previously predicted actions and collides with the hand.

\subsubsection{Place onto a Moving Target}

\begin{figure}[b]
    \centering
    \vspace{-14pt}
    \includegraphics[width=0.9\linewidth]{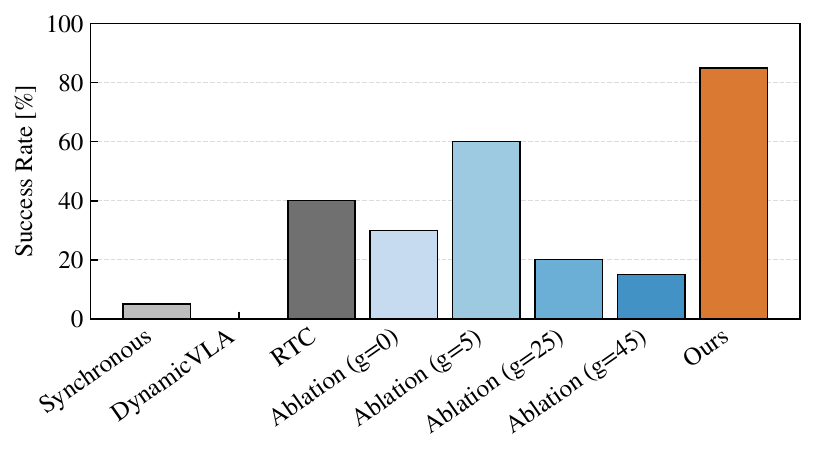}
    \caption{Success rates for Task 2 (\emph{Place onto a Moving Target}) across baseline methods, fixed-gap ablations, and our method.}
    \label{fig:task2_success}
\end{figure}

Unlike the preceding task, this task contains no picking phase and directly evaluates whether the robot can adapt its placement motion to a continuously moving target. As shown in Fig.~\ref{fig:task2_success}, synchronous execution and the fixed-gap ablation ($g=45$) achieve success rates of only 5\% and 15\%, respectively. These results indicate that infrequent policy updates provide insufficient responsiveness to the target motion.

Meanwhile, naively reducing the inference gap is not sufficient. The ablation with $g=0$ achieves a success rate of only 30\%, substantially lower than the 60\% achieved with $g=5$. By dynamically adapting the inference gap to the observed target motion, our method achieves the highest success rate of 85\%, outperforming RTC by 45 percentage points and the best fixed-gap ablation by 25 percentage points.

\begin{figure}[h]
    \centering
    \includegraphics[width=\linewidth]{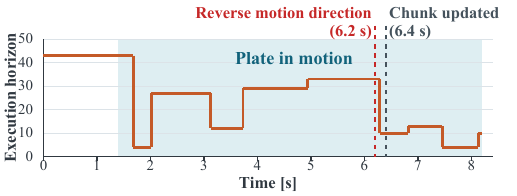}
    \caption{Execution horizon during Task~2 under target perturbation.
The shaded region indicates the plate in motion; dashed lines mark motion reversal and the subsequent chunk update.}
    \label{fig:moving-target}
\end{figure}

Fig.~\ref{fig:task2-process} illustrates a representative execution of our method during moving-target placement. At $t=6.2$\,s, the plate is pushed toward the robot base, and the event estimator detects this change and triggers a new round of policy inference. The updated action chunk becomes available and begins execution at \(t=6.4\,\mathrm{s}\), resulting in the shortened execution horizon shown in Fig.~\ref{fig:moving-target}. As visualized in Fig.~\ref{fig:task2-process}, the updated chunk, shown in blue, redirects the robot toward the new plate position, whereas the unexecuted suffix of the previous chunk, shown in orange, continues toward the outdated target position. This example demonstrates how adaptive inference timing replaces outdated actions when needed, enabling prompt responses without relying on continuously high-frequency inference.

\subsection{Overall Performance Across Evaluation Settings}

\begin{figure}[h]
    \centering
    \includegraphics[width=0.9\columnwidth]{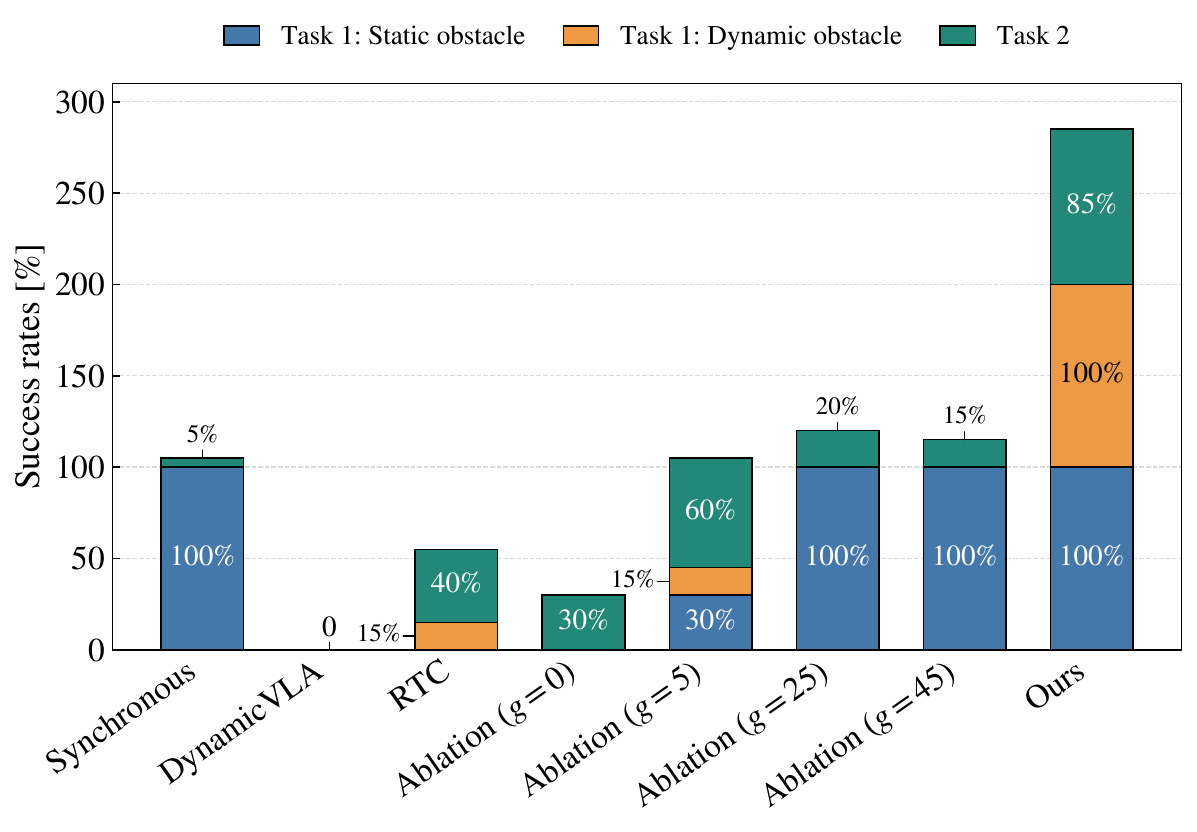}
    \caption{Aggregate full-task success across three evaluation settings, with each stacked segment representing one setting.}
    \label{fig:overall_success}
\end{figure}

To evaluate the overall robustness of the execution methods, Fig.~\ref{fig:overall_success} aggregates the full-task success rates across the three evaluation settings: Task~1 with a static obstacle (full-horizon), Task~1 with a dynamic obstacle (full-horizon), and Task~2. We compute the overall performance as the sum of the full-task success rates across the three equally weighted evaluation settings.

The baseline methods exhibit clear scenario-dependent trade-offs. Synchronous execution and the long-gap ablations perform reliably in the static setting but fail to respond effectively to dynamic changes. Conversely, RTC and the short-gap ablations provide greater responsiveness in dynamic settings but compromise execution reliability during phases in which frequent action-chunk merging is unnecessary. This trade-off becomes particularly evident in the full-task evaluation of Task~1 with a dynamic obstacle, which combines stable manipulation phases with dynamic obstacle avoidance. Consequently, all baseline methods achieve full-task success rates of no more than 15\%.

In contrast, our method dynamically adapts the inference timing according to the detected scene events. It maintains a long inference gap during stable phases to preserve motion consistency, while promptly triggering new policy inference when significant environmental changes are detected. This adaptive behavior enables reliable execution in static settings while retaining sufficient responsiveness to moving obstacles and targets. Consequently, our method achieves success rates of 100\%, 100\%, and 85\% across the three settings, demonstrating consistently high task performance under both static and dynamic conditions.

\section{Conclusion}

This paper presented an event-triggered dynamic inference strategy for responsive and reliable asynchronous VLA execution. Unlike existing methods that use a fixed inference gap, our approach adapts inference timing according to scene changes observed since the most recent policy inference. It maintains the largest feasible gap when no significant event is detected and triggers early inference upon event detection. Across real-world settings spanning static and dynamic conditions, our method achieves the highest full-task success rate in every setting, outperforming the strongest baseline by $55$ percentage points. These results demonstrate that dynamic inference scheduling can simultaneously preserve motion continuity and enable prompt reactions to environmental changes. Future work will evaluate the proposed strategy across a broader range of manipulation scenarios and VLA backbones.

\section*{ACKNOWLEDGMENT}
The authors used GPT-5.6 to assist with grammatical and linguistic refinement, as they are non-native English speakers. Additionally, GPT-5.6 was utilized to help format and refine the visual presentation of selected figures. Codex was used to help clean the codebase for its open-source release. All generated suggestions and modifications were reviewed and verified by the authors.

\bibliography{IEEEabrv,mybib2022}
\bibliographystyle{myIEEEtran}

\label{last-page}
\end{document}